%% file: deinterlace.tex
\documentclass[acmtog]{acmart}

\usepackage{booktabs} 
\usepackage[ruled]{algorithm2e} 
\usepackage{multirow}

\DeclareMathOperator*{\argmin}{\arg\min}

\SetAlFnt{\small}
\SetAlCapFnt{\small}
\SetAlCapNameFnt{\small}
\SetAlCapHSkip{0pt}
\IncMargin{-\parindent}
\renewcommand\footnotetextcopyrightpermission[1]{} 
\acmJournal{TOG}
\acmVolume{X}
\acmNumber{X}
\acmArticle{XX}
\acmYear{X}
\acmMonth{8}

\setcopyright{none}

\acmDOI{0000001.0000001_2}

\begin{document}
\title{Real-time Deep Video Deinterlacing}

\author{Haichao Zhu}
\orcid{1234-5678-9012}
\affiliation{%
  \institution{The Chinese University of Hong Kong}
}
\email{hczhu@cse.cuhk.edu.hk}

\author{Xueting Liu}
\affiliation{%
  \institution{The Chinese University of Hong Kong}
}
\email{xtliu@cse.cuhk.edu.hk}

\author{Xiangyu Mao}
\affiliation{%
	\institution{The Chinese University of Hong Kong}
}
\email{maoxiangyu@sensetime.com}

\author{Tien-Tsin Wong}
\affiliation{%
	\institution{The Chinese University of Hong Kong}
}
\email{ttwong@cse.cuhk.edu.hk}


\begin{abstract}
\input{abstract.tex}
\end{abstract}

\ccsdesc[500]{Computing methodologies~Reconstruction}
\ccsdesc[500]{Computing methodologies~Neural networks}

%
%


\keywords{Video deinterlace, image interpolation, convolutional neural network, deep learning}

\begin{teaserfigure}
  \includegraphics[width=\textwidth]{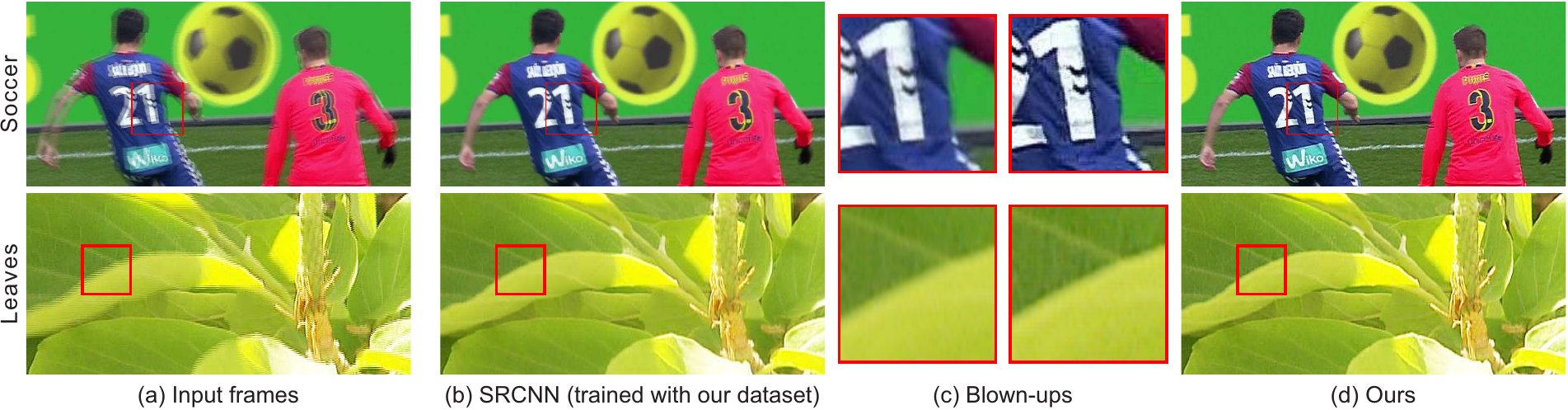}
  \caption{(a) Input interlaced frames. (b) Deinterlaced results generated by SRCNN~\cite{dong2016image} re-trained with our dataset. (c) Blown-ups from (b) and (d) respectively. (d) Deinterlaced results generated by our method. The classical super-resolution method SRCNN reconstruct each frame based on a single field and has large information loss. It also follows the conventional translation-invariant assumption which does not hold for the deinterlacing problem. Therefore, it inevitably generates blurry edges and artifacts, especially around sharp boundaries. In contrast, our method can circumvent this issue and reconstruct frames with higher visual quality and reconstruction accuracy.}
  \label{fig:teaser}
\end{teaserfigure}

\maketitle

\section{Introduction}
\input{introduction.tex}

\section{Related Work}
\input{relatedWorks.tex}

\section{Overview}
\input{overview.tex}
\section{DCNN-based Video Deinterlacing}\label{sec:deinterlacing}
\input{method.tex}
\section{Result and Discussion}
\input{experiment.tex}
\section{Conclusion}
\input{conclusion.tex}

\citestyle{acmauthoryear}
\setcitestyle{square}
\bibliographystyle{ACM-Reference-Format}
\bibliography{papers}

\end{document}

%% file: abstract.tex
Interlacing is a widely used technique, for television broadcast and video
recording, to double the perceived frame rate without increasing the bandwidth.
But it presents annoying visual artifacts, such as flickering and silhouette
``serration,''  during the playback. Existing state-of-the-art deinterlacing
methods either ignore the temporal information to provide real-time performance
but lower visual quality, or estimate the motion for better deinterlacing but
with a trade-off of higher computational cost. In this paper, we present 
the first and novel deep convolutional neural networks (DCNNs) based method to
deinterlace with high visual quality and real-time performance. Unlike existing
models for super-resolution problems which relies on the translation-invariant
assumption, our proposed DCNN model utilizes the temporal information from both
the odd and even half frames to reconstruct only the missing scanlines, and retains
the given odd and even scanlines for producing the full deinterlaced frames.
By further introducing a layer-sharable architecture, our system can achieve
real-time performance on a single GPU. Experiments shows that our
method outperforms all existing methods, in terms of 
reconstruction accuracy and computational performance.

%% file: introduction.tex

\begin{figure}[!tp]	
	\includegraphics[width=1\linewidth]{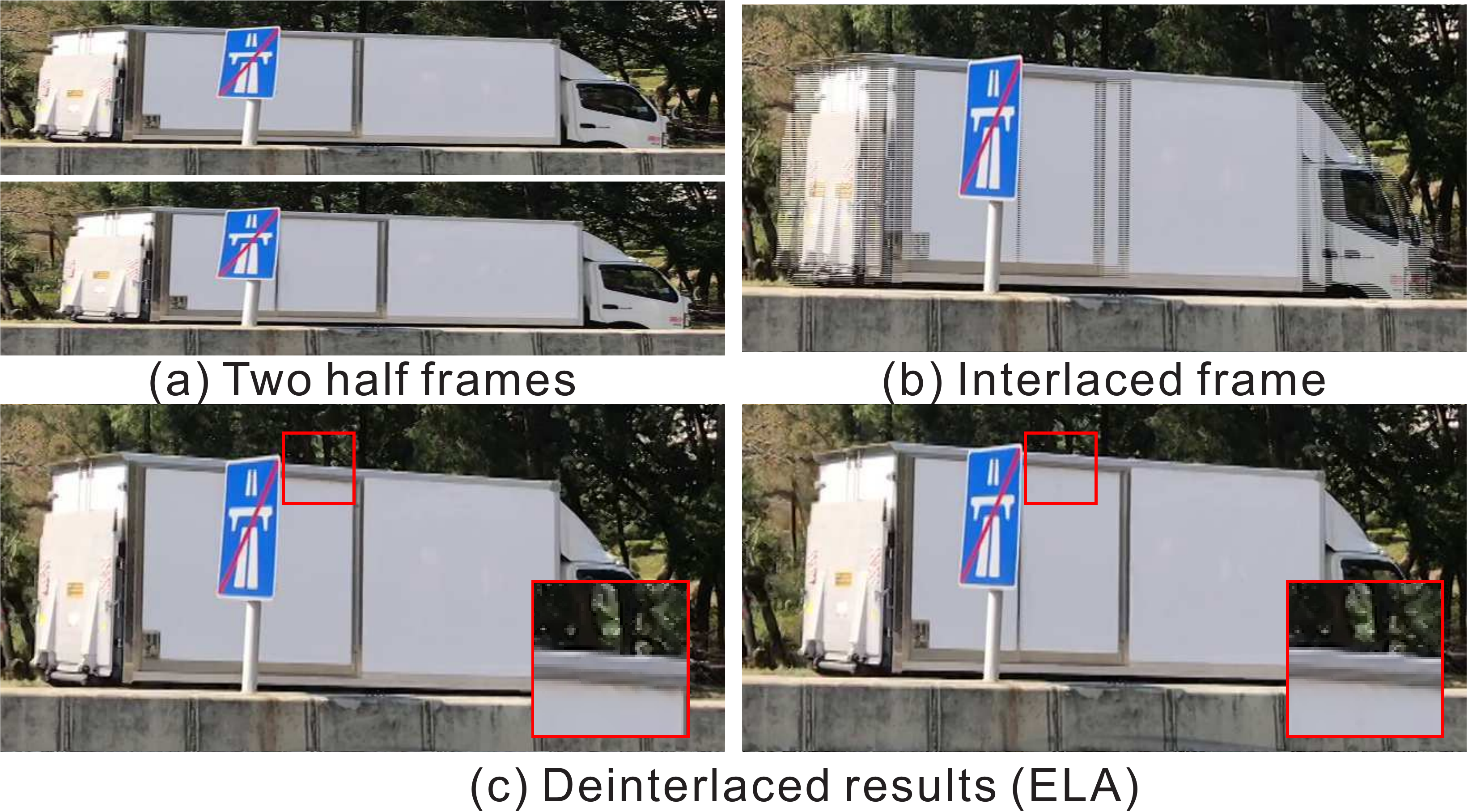}\\
	\caption{(a) Two half fields are captured in two distinct time instances. 
	(b) The interlaced display exhibits obvious artifacts on the silhouette of moving car.
	(c) Two full frames	reconstructed from the two half frames independently with an intra-field deinterlacing method ELA~\cite{doyle1990interlaced}.} \label{fig:interlaced_example}
\end{figure}

Interlacing technique has been widely used in the past few decades for television
broadcast and video recording, in both analog and digital ways. Instead of capturing all
$N$ scanlines for each frame, only $N/2$ odd numbered scanlines are captured for
the current frame (Fig.~\ref{fig:interlaced_example}(a), upper), and the other
$N/2$ even numbered scanlines are captured for the following frame
(Fig.~\ref{fig:interlaced_example}(a), lower). It basically trades the frame
resolution for the frame rate, in order to double the perceived frame rate
without increasing the bandwidth. Unfortunately, since the two half frames are
captured in {\em different time instances}, there are significant visual
artifacts such as line flickering and ``serration'' on the silhouette of moving
objects (Fig.~\ref{fig:interlaced_example}(b)), when the odd and even fields are
interlaced displayed. The degree of ``serration'' depends on the motion of objects
and hence is spatially varying. This makes deinterlacing (removal of interlacing artifacts) an ill-posed problem.



Many deinterlacing methods have been proposed to suppress the visual artifacts.
A typical approach is to reconstruct two full frames from the odd and even half
frames independently (Fig.~\ref{fig:interlaced_example}(c)). However, the result
is usually unsatisfactory, due to the large information 
loss (50\% loss)~\cite{doyle1990interlaced,wang2012efficient,wang2013moving}. 
Higher-quality reconstruction can be obtained by first estimating object 
motion~\cite{jeon2009weighted,mohammadi2012enhanced,lee2013high}. However, motion
estimation from half interlacing frames are not reliable, and also computationally expensive.
Hence, they are seldomly used in practice, 
let alone real-time applications.

\begin{figure}[!tp]
\includegraphics[width=1\linewidth]{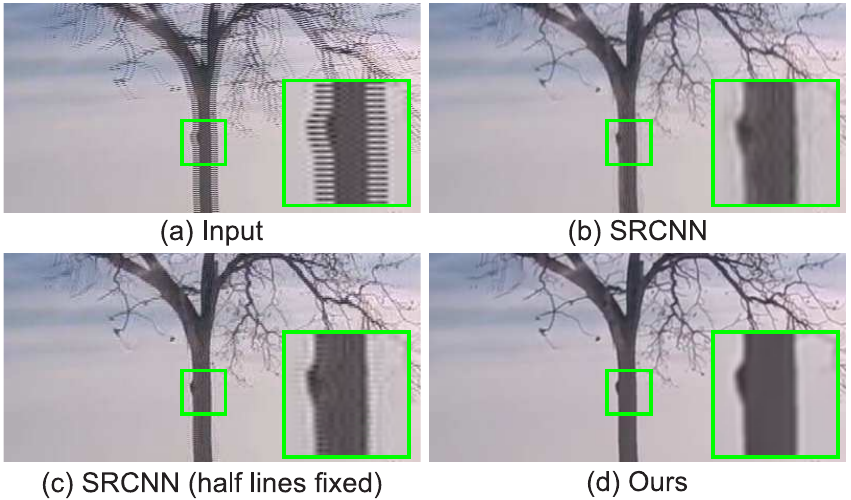}\\
\caption{(a) An input interlaced frame. 
(b) Directly applying SRCNN to deinterlacing introduces blurry and halo artifacts. 
(c) The visual artifacts are worsen if we retain the pixels from the input odd/even scanlines. 
(d) Our result.}\label{fig:srcnn_problem} 
\end{figure}

In this paper, we propose the first deep convolutional neural networks (DCNNs)
method tailormade for the video deinterlacing problem. To our best knowledge, no
DCNN-based deinterlacing method exists. One may argue that existing DCNN-based
methods for interpolation or super-resolution~\cite{mallat2016understanding,dong2016image} can be applied to
reconstruct the full frames from the half frames, in order to solve the
deinterlacing problem. However, such naive approach lacks of utilizing the
temporal information between the odd and even half frames, just like the existing
intra-field deinterlacing methods~\cite{doyle1990interlaced,wang2012efficient}. Moreover,
this naive approach follows the conventional translation-invariant assumption.
That means, all pixels in the output full frames are processed with the same set
of convolutional filters, even though half of the scanlines (odd/even numbered)
actually exist in the input half frames. Fig.~\ref{fig:srcnn_problem}(b) shows
a full frame, reconstructed by the state-of-the-art DCNN-based super-resolution method,
SRCNN~\cite{dong2016image}, exhibiting obvious halo artifact. 
Instead of replacing the potentially error-contaminated pixels from the convolutional filtering with the groundtruth
pixels in the input half frames and leading to visual artifacts
(Fig.~\ref{fig:srcnn_problem}(c)), we argue that we should only reconstruct the
missing scanlines, and leave the pixels in the original odd/even scanlines
intact. All these motivate us to design a novel DCNN model tailored for
solving the deinterlacing problem.

In particular, our newly proposed DCNN architecture circumvents the translation-invariant assumption and takes the temporal
information into consideration. Firstly, we only estimate the missing scanlines
to avoid modifying the groundtruth pixel values from the odd/even scanlines
(input). That is, the output of the neural network system are two half frames
containing only the missing scanlines. Unlike most existing methods which ignore the
temporal information between the odd and even frames, we reconstruct each half 
output frame from both the odd and even frames. In other words,
 our
neural network system takes two original half frames as input and outputs two missing half
frames (complements). 

Since we have two outputs, two neural networks are needed for training. 
We further accelerate it by combining the lower-levels of two
neural networks~\cite{bengio2012deep}, as the input are the same and 
hence the lower-level convolutional filters are sharable. 
With this improved network structure, we can achieve real-time performance.

To validate our method, we evaluate it over a rich
variety of challenging interlaced videos including live broadcast, legacy movies,
and legacy cartoons. Convincing and visually pleasant results are obtained in all experiments 
(Fig.~\ref{fig:teaser} \& \ref{fig:srcnn_problem}(d)). We also compare our
method to existing deinterlacing methods and DCNN-based models in both visual comparison and quantitative
measurements. All experiments confirm that our method not
only outperforms existing methods in terms of accuracy, but also
speed performance. 



%% file: relatedWorks.tex
Before introducing our method, we first review existing works related to 
deinterlacing. They can be roughly classified into tailor-made 
deinterlacing methods,  traditional image resizing methods, and DCNN-based
image restoration approaches.

\vspace{0.15in}
\noindent\emph{Image/Video Deinterlacing}\,\,\,\,
\begin{figure*}[!tp]
  \includegraphics[width=1\linewidth]{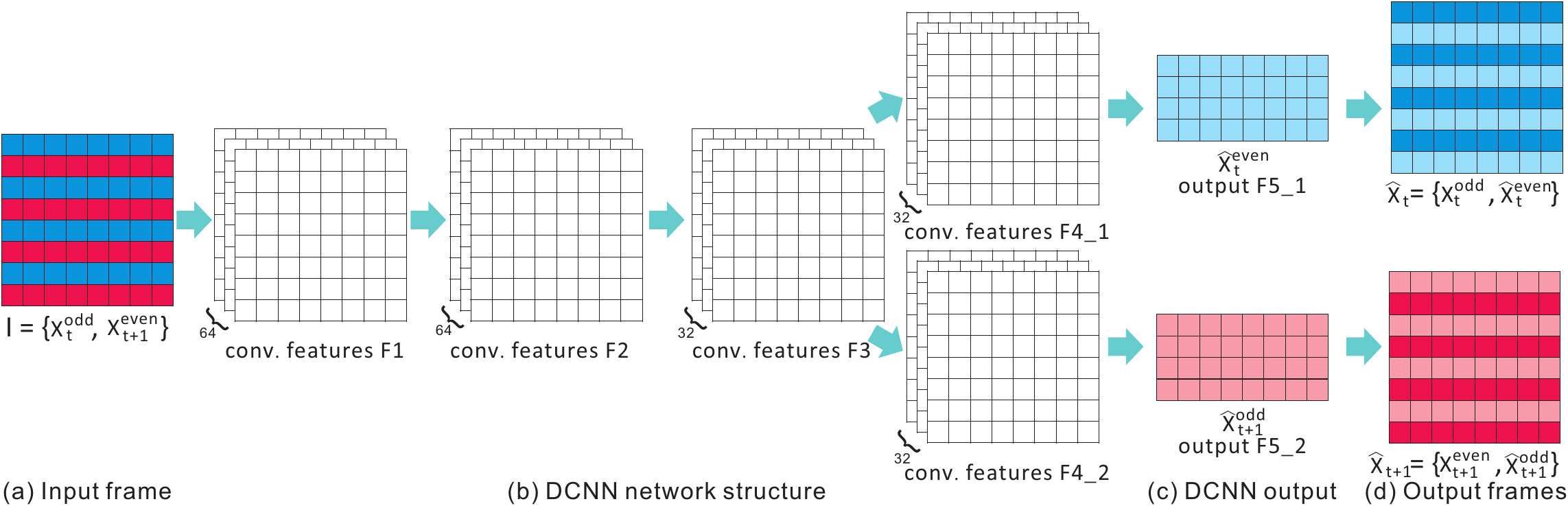}\\
  \caption{The architecture of the proposed convolutional neural network.}\label{fig:cnn_model}
\end{figure*}
Image/video deinterlacing is a classic  vision problem. 
Existing methods can be classified into two categories: intra-field deinterlacing
\cite{doyle1990interlaced,wang2012efficient,wang2013moving} and inter-field
deinterlacing \cite{jeon2009weighted,mohammadi2012enhanced,lee2013high}. 
Intra-field deinterlacing methods reconstruct two full frames from the odd and even fields
independently. Since
there is large information loss (half of the data is missing) during frame
reconstruction, the visual quality is usually less satisfying. To improve visual
quality, inter-field deinterlacing methods incorporate the temporal
information between multiple fields from neighboring frames during frame
reconstruction. Accurate motion compensation or motion
estimation~\cite{horn1981determining} is needed to achieve satisfactory quality. However, 
accurate motion estimation is hard in general. In
addition, motion estimation requires high computational cost, and hence inter-field
deinterlacing methods are seldom used in practice, especially for applications
requiring real-time processing.

\vspace{0.15in}
\noindent\emph{Traditional Image Resizing}\,\,\,\,
Traditional image resizing
methods can also be used for deinterlacing by scaling up the height of each field.
To scale up an image, cubic~\cite{mitchell1988reconstruction} and
Lanczos interpolation~\cite{duchon1979lanczos} are frequently used. While they work well for low-frequency components, high-frequency components 
(e.g. edges) may be over-blurred. More advanced image resizing methods, 
such as kernel regression~\cite{takeda2007kernel} and
bilateral filter~\cite{hung2012fast} can improve the
visual quality by preserving more high-frequency components. 
However, these methods may still introduce noise or artifacts 
if the vertical sampling rate is less than the Nyquist rate. 
More critically, they only utlize a single field and ignore the temporal information, 
and hence suffer the same problem as intra-deinterlacing methods.


\vspace{0.15in}

\noindent\emph{DCNNs for Image Restoration}\,\,\,\, In recent years, 
deep convolutional neural networks (DCNNs) based methods have been proposed to solve 
many image restoration problems.
Xie et al.~\shortcite{xie2012image} proposed a DCNN model for
image denosing and inpainting. This model recovers the values of corrupted
pixels (or missing pixels) by learning the mapping between corrupted and
uncorrupted patches. Dong et al.~\shortcite{dong2016image} proposed to adopt
DCNN for image super-resolution, which greatly outperforms the state-of-the-art
image super-resolution methods. Gharbi et al.~\shortcite{gharbi2016deep} further
proposed a DCNN model for joint demosaiking and denosing. It infers
the values of three color channels of each pixel from a single noisy
measurement.


It seems that we can simply re-train these state-of-the-art 
neural network based methods for our deinterlacing purpose. 
However, our experiments show that visual artifacts are still unavoidable, as 
these DCNNs generally follow the conventional translation-invariant assumption and
modify the values of all pixels, even in the known odd/even scanlines.  Using a larger
training dataset or deeper network structure may alleviate this problem, but the
computational cost is drastically increased and still there is no guarantee 
that the values of the known pixels remain intact. 
Even if we fix the values of the known pixels (Fig.~\ref{fig:srcnn_problem}(c)), 
the quality does not improve. 
In contrast, we propose a novel DCNN tailored for deinterlacing.
Our model only estimates the missing pixels instead of the whole frame, and also
take the temporal information into account to improve visual quality.

%% file: overview.tex
Given an input interlaced frame $\mathbf{I}$ (Fig.~\ref{fig:cnn_model}(a)),  our
goal of deinterlacing is to reconstruct two full size original frames
$\mathbf{X}_t$ and $\mathbf{X}_{t+1}$ from $\mathbf{I}$
(Fig.~\ref{fig:cnn_model}(d)). We denote the odd field of $\mathbf{I}$ as
$\mathbf{X}^{\text{odd}}_t$ (blue pixels in Fig.~\ref{fig:cnn_model}(a)), and
the even field of $\mathbf{I}$ as $\mathbf{X}^{\text{even}}_{t+1}$ (red pixels in
Fig.~\ref{fig:cnn_model}(a)). The superscripts, $\text{odd}$ and $\text{even}$,
denote the odd- or even-numbered half frames. The subscripts, $t$ and $t+1$,
denote the two fields are captured at two different time instances. Our goal is
to reconstruct two missing half frames, $\mathbf{X}^{\text{even}}_t$ (light blue
pixels in Fig.~\ref{fig:cnn_model}(c)) and $\mathbf{X}^{\text{odd}}_{t+1}$ (pink
pixels in Fig.~\ref{fig:cnn_model}(c)). Note that we retain the known fields
$\mathbf{X}^{\text{odd}}_t$  (blue pixels) and $\mathbf{X}^{\text{even}}_{t+1}$
(red pixels) in our two output full frames (Fig.~\ref{fig:cnn_model}(d)).

To estimate the unknown pixels $\mathbf{X}^{\text{even}}_t$ and
$\mathbf{X}^{\text{odd}}_{t+1}$ from the interlaced frame $\mathbf{I}$, we
propose a novel DCNN model (Fig.~\ref{fig:cnn_model}(b) \& (c)). The input
interlaced frame can be of any resolution, and two half output images are
obtained with five convolutional layers. The weights of the convolutional
operators are trained from a DCNN model training procedure based on a prepared
training dataset. During the training phase, we synthesize a set of interlaced
videos from progressive videos of different types as the training pairs. The
reason that we need to synthesize interlaced videos for training is that no
groundtruth exists for the existing interlaced videos captured by interlaced
scan devices. The details of preparing the training dataset and the design
of the proposed DCNN are described in Section~\ref{sec:deinterlacing}. 

%% file: method.tex
\subsection{Training Data Preparation}

\begin{figure}[!tp]
	\includegraphics[width=1\linewidth]{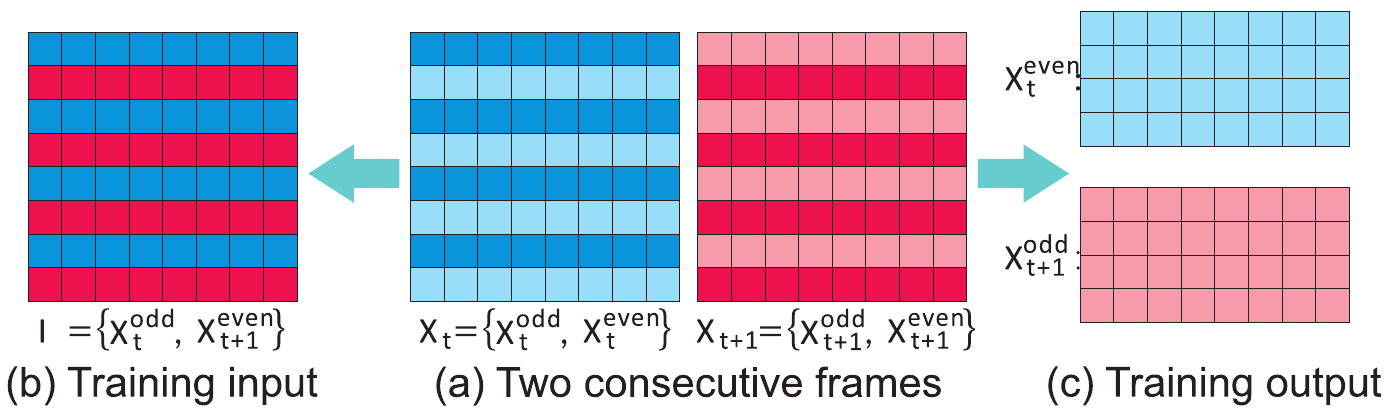}\\
	\caption{Training data preparation. (a) Two consecutive frames $\mathbf{X}_t$ and $\mathbf{X}_{t+1}$ from an input video. (a) An interlaced frame $\mathbf{I}$ is synthesized by taking the odd lines from $\mathbf{X}_t$ and even lines from $\mathbf{X}_{t+1}$ respectively and regarded as the training input. (c) The even lines of $\mathbf{X}_t$ and the odd lines of $\mathbf{X}_{t+1}$ are regarded as the training output.}\label{fig:data_preparing}
\end{figure}

\begin{figure}[!tp]
	\includegraphics[width=1\linewidth]{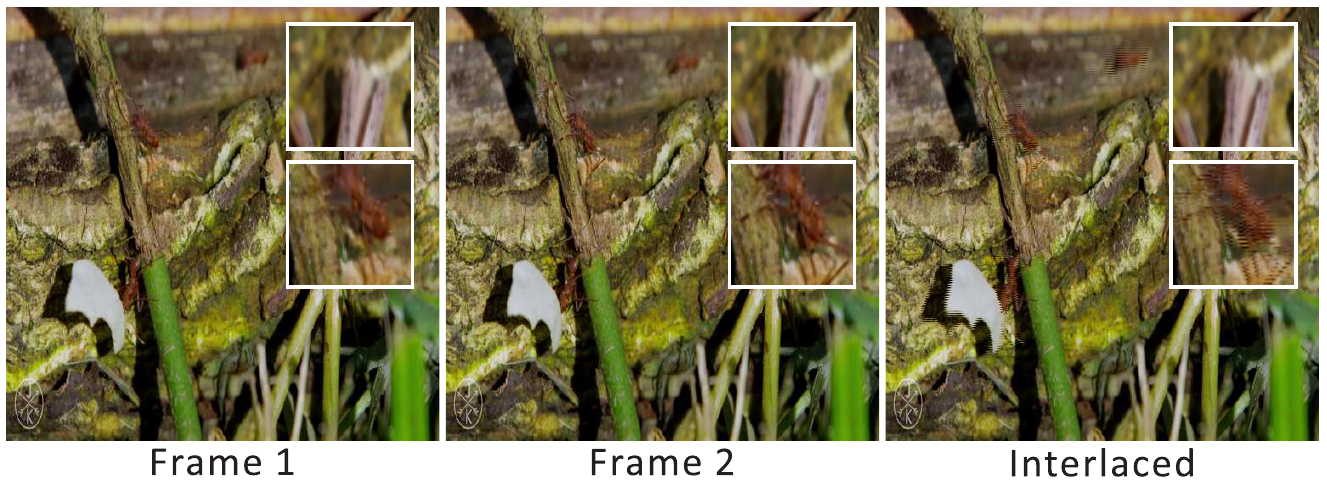}\\
	\caption{A real example of synthesizing an interlaced frame from two consecutive progressive frames.}\label{fig:data_preparing_real}
\end{figure}

While there exists a large collection of interlaced videos over the Internet,
unfortunately, the ground-truth of these videos is lacking. Therefore, to
prepare a training data set, we have to synthesize
interlaced videos from existing progressive videos. To enrich our data variety, we
collect $33$ videos from the Internet and capture $18$ videos using progressive
scan devices ourselves. The videos are of different genres, ranging from 
scenic, sports, computer-rendered, to classic movies and cartoons. Then we randomly sample $3$ pairs of consecutive frames from each collected
video and obtain $153$ frame pairs in total. For each pair of consecutive
frames, we rescale each frame to the size of $512\times512$ and label them as the
pair of original frames $\mathbf{X}_t$ and $\mathbf{X}_{t+1}$ (ground-truth full frames)
(Fig.~\ref{fig:data_preparing}(a)). Then we synthesize an interlaced frame based
on these two original frames as
$\mathbf{I}=\{\mathbf{X}^{\text{odd}}_t,\mathbf{X}^{\text{even}}_{t+1}\}$, i.e.,
the odd lines of $\mathbf{I}$ are copied from $\mathbf{X}_t$ and the even lines
of $\mathbf{I}$ are copied from $\mathbf{X}_{t+1}$
(Fig.~\ref{fig:data_preparing}(b) \&~\ref{fig:data_preparing_real}). For
each triplet $\left\langle \mathbf{I}, \mathbf{X}_t,
\mathbf{X}_{t+1}\right\rangle $ of $512\times512$ resolution, we further divide
them into $64\times64$-resolution patch triplets $\left\langle \mathbf{I}_{p},
\mathbf{X}_{t,p}, \mathbf{X}_{t+1,p}\right\rangle$ with the sampling stride
setting to $64$.  Note that during patch generation, the parity of the divided
patches remain the same as original images. Finally, for each patch triplet
$\left\langle \mathbf{I}_{p}, \mathbf{X}_{t,p}, \mathbf{X}_{t+1,p}\right\rangle
$, we use $\mathbf{I}_{p}$ as a training input
(Fig.~\ref{fig:data_preparing}(b)) and the corresponding
$\mathbf{X}^{\text{even}}_{t,p}$ and $\mathbf{X}^{\text{odd}}_{t+1,p}$ as
training outputs (Fig.~\ref{fig:data_preparing}(c)). In particular, we convert
patches into the Lab color space and only use the L channel for training. Altogether, we
collect 9,792 patch triplets from the prepared videos, where $80\%$ of
the triplets are used for training and the rest are used for validation during
the training process. Note that, although our model is trained by patches of
$64\times64$ resolution, the trained convolutional operators can actually be
applied on images of any resolution.

\subsection{Neural Network Architecture}

With the prepared training dataset, we now present how we design our network
structure for deinterlacing. An illustration of our network structure is shown
in Fig.~\ref{fig:cnn_model}. It contains five convolutional layers. Our goal is
to reconstruct the original two frames $\mathbf{X}_t$ and $\mathbf{X}_{t+1}$
from an input interlaced frame $\mathbf{I}$. In the following, we first explain
our  design rationales and then describe the architecture in
detail.

\begin{figure}[!tp]
	\centering
	\includegraphics[width=\linewidth]{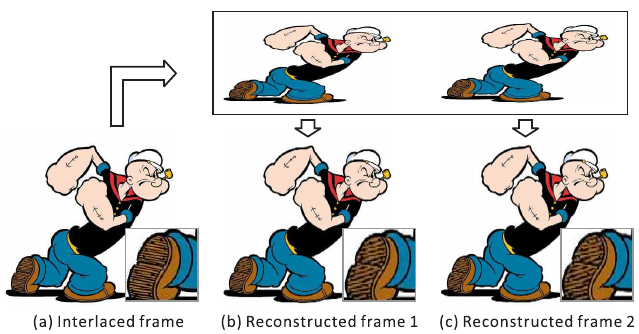}\\
	\caption{Reconstructing two frames from two fields independently leads to inevitable visual artifacts due to the large information loss.}\label{fig:input_problem}
\end{figure}

\vspace{0.15in}
\noindent\emph{The Input/Output Layers}\,\,\,\,
One may suggest to utilize the existing neural network (e.g. SRCNN
\cite{dong2016image}) to learn $\mathbf{X}_t$ from
$\mathbf{X}^{\text{odd}}_{t}$ and $\mathbf{X}_{t+1}$ from
$\mathbf{X}^{\text{even}}_{t+1}$ independently. This effectively turns the problem into 
a super-resolution or image upscaling problem.
However, there are two drawbacks.

First of all, since the two frame reconstruction processes (i.e. from
$\mathbf{X}^{\text{odd}}_{t}$ to $\mathbf{X}_t$ and
$\mathbf{X}^{\text{even}}_{t+1}$ to $\mathbf{X}_{t+1}$) are independent from
each other, the neural network can only estimate the full frame from the known
half frame without the temporal information. This inevitably leads to less
satisfying results due to the large (50\%) information loss. In fact, the two
fields in the interlaced frame are temporally correlated. Consider an extreme
case where the scene in the two consecutive frames are static. In this scenario,
the two consecutive frames are exactly the same, and the interlaced frame should
also be artifact-free and exactly equal to the groundtruth we are looking for.
However, using this naive super-resolution approach, we have to feed the half
frame $\mathbf{X}^{\text{odd}}_t$ (or $\mathbf{X}^{\text{even}}_{t+1}$) to
reconstruct a full frame. It completely ignores the another half frame (which
now contains the exact pixel values) and introduces artifacts (due to 50\%
information loss). Fig.~\ref{fig:input_problem} shows the  poor result of one
such scenario. In contrast, our proposed neural network takes the whole
interlaced frame $\mathbf{I}$ as input (Fig.~\ref{fig:cnn_model}(a)). Note that
the temporal information is implicitly taken into consideration in our network,
since the two fields captured at different time instances are used for reconstructing each
single frame. The network may exploit the temporal correlation between fields to
improve the visual quality in higher-level convolutional layers.

Secondly, the standard neural network generally follows the conventional
translation-invariant assumption. That means all pixels in the input image are
processed with the same set of convolutional filters. However, in our deinterlacing
application, half of the pixels in $\mathbf{X}_t$ and $\mathbf{X}_{t+1}$ actually 
exist in $\mathbf{I}$ and should be directly copied from $\mathbf{I}$. 
Applying convolutional filters on these
known pixels inevitably changes their original colors and leads to clear
artifacts (Fig.~\ref{fig:srcnn_problem}(b) \& (c)). In contrast, our neural network only
estimates the unknown pixels $\mathbf{X}^{\text{even}}_t$ and
$\mathbf{X}^{\text{odd}}_{t+1}$ (Fig.~\ref{fig:cnn_model}(c)) and copies the known
pixels from $\mathbf{I}$ to $\mathbf{X}_t$ and $\mathbf{X}_{t+1}$ directly
(Fig.~\ref{fig:cnn_model}(d)).



\vspace{0.15in}
\noindent\emph{Pathway Design}\,\,\,\,
Since we estimate two half frames $\mathbf{X}^{\text{even}}_t$ and
$\mathbf{X}^{\text{odd}}_{t+1}$ from the interlaced frame $\mathbf{I}$, we
actually have to train two networks/pathways independently. Separately training two
networks is computational costly. Instead of training two networks, one
may suggest to train a single network for estimating the two half frames simultaneously
by doubling the depth of each convolutional layer. However, this also
highly increases the computational cost, since the number of the trained weights
are doubled. As reported by \cite{bengio2012deep}, deep neural network is to
seek good representation of input data, and such representations can be transferred
to many other tasks if the input data is similar. For example, the trained
features of AlexNet \cite{krizhevsky2012imagenet} (originally designed for
object recognition) can also be used for texture recognition and segmentation
\cite{cimpoi2015deep}. In fact, the lower-level layers of the convolutional
networks are always lower-level feature detectors that can detect edges and other
primitives. These lower-level layers in the trained models can be reused for new
tasks by training new higher-level layers on top of them. Therefore, in our
deinterlacing scenario, it is natural to combine the lower-level convolutional
layers to reduce the computation, since the input of the two networks/pathways is
completely the same. On top of these weight-sharing lower-level layers, higher-level 
layers are trained separately for estimating $\mathbf{X}^{\text{even}}_t$
and $\mathbf{X}^{\text{odd}}_{t+1}$ respectively. This makes the 
higher-level layers more adaptable to different objectives. Our method can be regarded
as training one neural network for estimating $\mathbf{X}^{\text{even}}_t$ and
then fixing the first three convolutional layers and re-training a second neural
network for estimating $\mathbf{X}^{\text{odd}}_{t+1}$.


\vspace{0.15in}
\noindent\emph{Detailed Architecture}\,\,\,\,
As illustrated in Fig.~\ref{fig:cnn_model}(b) \& (c), our network contains five convolutional layers with weights.
The first, second, and third layers are sequentially connected and shared by both pathways.
The first convolutional layer has $64$ kernels of size $3\times3\times1$.
The second convolutional layer has $64$ kernels of size $3\times3\times64$ and is connected to the output of the first layer.
The third convolutional layer has $64$ kernels of size $1\times1\times64$ and is connected to the output of the second layer.
The forth and fifth layers branch into two pathways without any connection between them.
The forth convolutional layer has $64$ kernels of size $3\times3\times64$ where each pathway has $32$ kernels.
The fifth convolutional has $2$ kernels of size $3\times3\times64$ where each pathway has $1$ kernel.
The activations for the first two layers are ReLU functions, while for the rest layers are identify functions.
The strides of convolution for the first four layers are $1$ pixel.
For the last layer, the horizontal stride remains $1$ pixel, while the vertical stride is $2$ pixels to obtain half-height images.

\subsection{Learning and Optimization}
Given the training dataset containing a set of triplets $\left\langle \mathbf{I}_p,\mathbf{X}^{\text{even}}_{t,p},\mathbf{X}^{\text{odd}}_{t+1,p}\right\rangle$, the optimal weights $W^*$ of our neural network are trained via the following objective function:
\begin{equation}\label{eq:objective}
\begin{split}
W^*=\argmin \frac{1}{N_p} &\sum_p \Big(\|\widehat{\mathbf{X}}^{\text{even}}_{t,p}-\mathbf{X}^{\text{even}}_{t,p}\|^2_2+\|\widehat{\mathbf{X}}^{\text{odd}}_{t+1,p}-\mathbf{X}^{\text{odd}}_{t+1,p}\|^2_2\\
&+\lambda_{TV}\big(TV(\widehat{\mathbf{X}}_{t,p})+TV(\widehat{\mathbf{X}}_{t+1,p})\big)\Big)
\end{split}
\end{equation}
where $N_p$ is the number of training samples, $\widehat{\mathbf{X}}^{\text{even}}_{t,p}$ and $\widehat{\mathbf{X}}^{\text{odd}}_{t+1,p}$ are the estimated output of the neural network, $TV(\cdot)$ is the total variation regularizer~\cite{aly2005image,johnson2016perceptual} and $\lambda_{TV}$ is the regularization scalar.


We trained our neural network using Tensorflow on a workstation equipped with a
single nVidia TITAN X Maxwell GPU. The standard ADAM
optimization method \cite{kingma2014adam} is used to solve
Eq.~\ref{eq:objective}. The learning rate is $0.001$ and $\lambda_{TV}$ is set to
$2\times10^{-8}$ in our experiments. The number of epochs is $200$ and the batch size for each
epoch is $64$. It takes about $4$ hours to train the neural network.

%% file: experiment.tex
\begin{figure*}[!tp]
	\centering
	\includegraphics[width=\linewidth]{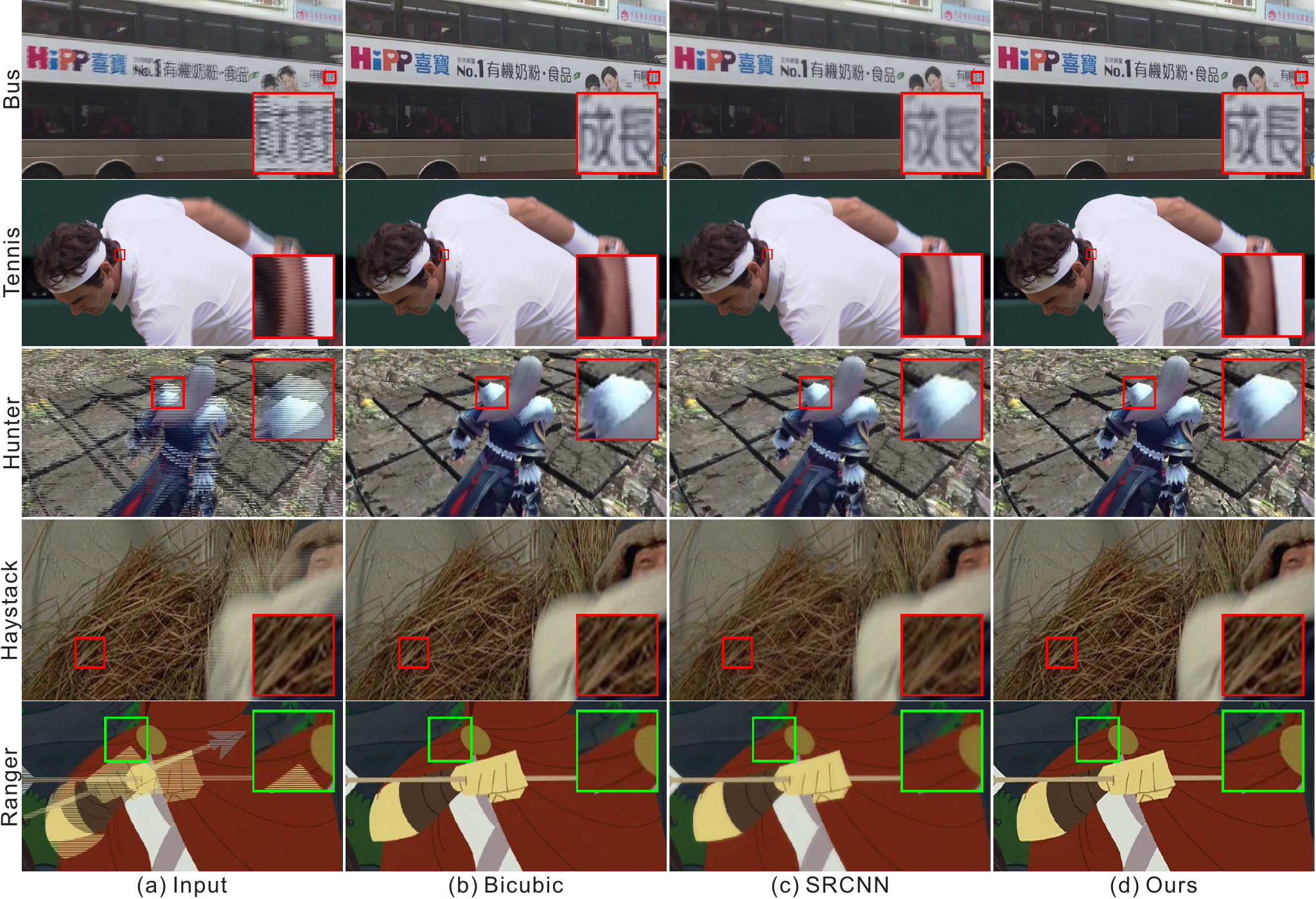}\\
	\caption{Comparisons between bicubic interpolation, SRCNN~\cite{dong2016image} and our method. 
	}\label{fig:compared_with_bicubic_and_srcnn_results}
\end{figure*}

\begin{figure*}[!tp]
	\centering
	\includegraphics[width=\linewidth]{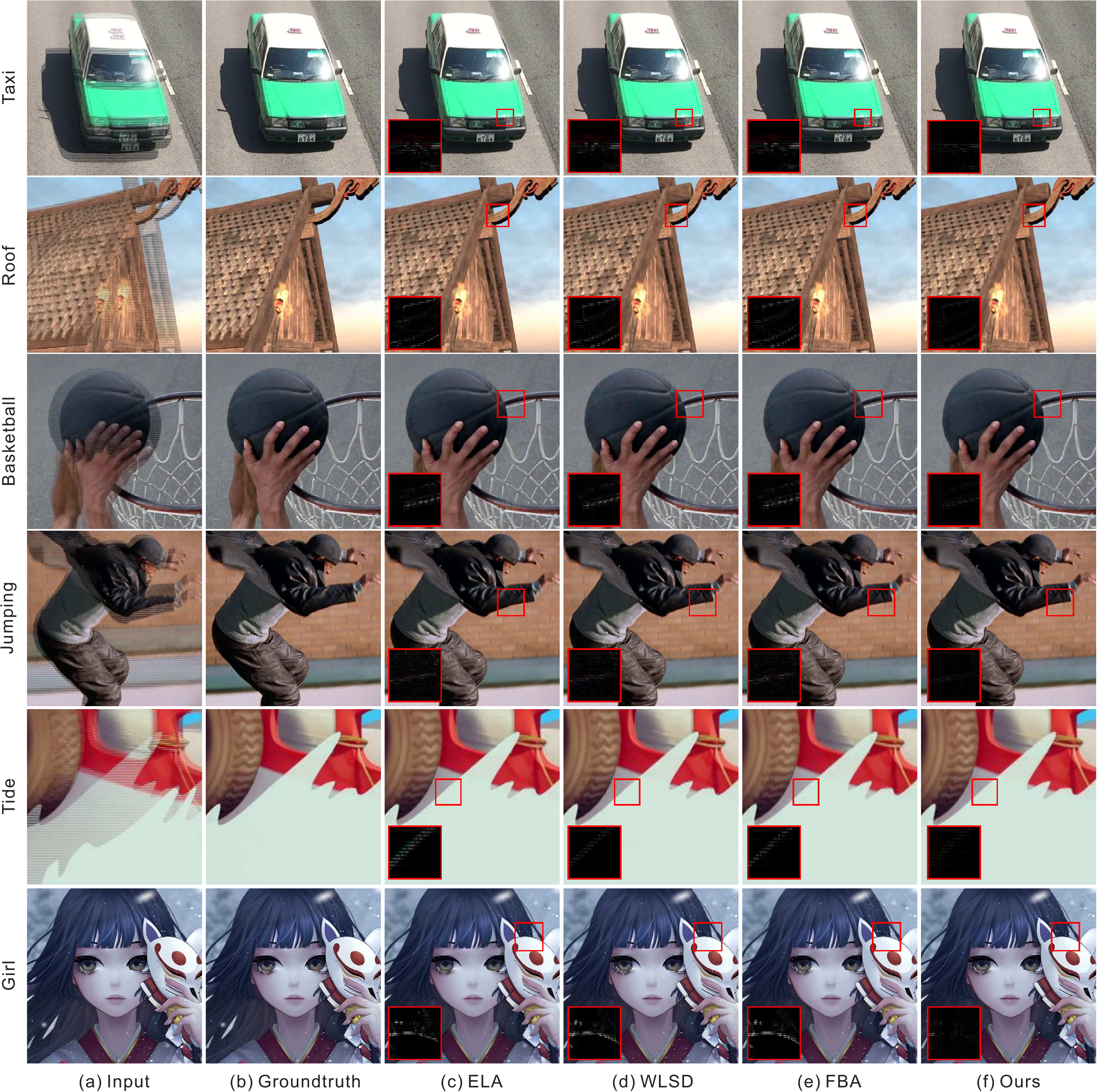}\\
	\caption{Comparisons between the state-of-the-art deinterlacing tailored methods, including ELA~\cite{doyle1990interlaced}, WLSD~\cite{wang2014interlacing}, and FBA~\cite{vedadi2013interlacing}, with our method. }\label{fig:compared_with_deinterlaced_method}
\end{figure*}

We evaluate our method on a large collection of interlaced videos downloaded
from the Internet or captured by ourselves with interlaced scan cameras. These
videos include live sporting videos (``Soccer'' in
Fig.~\ref{fig:teaser} and ``Tennis'' in
Fig.~\ref{fig:compared_with_bicubic_and_srcnn_results}), scenic videos
(``Leaves'' in Fig.~\ref{fig:teaser} and ``Bus'' in
Fig.~\ref{fig:compared_with_bicubic_and_srcnn_results}), computer-rendered gameplay
videos (``Hunter'' in Fig.~\ref{fig:compared_with_bicubic_and_srcnn_results}),
legacy movies (``Haystack'' in
Fig.~\ref{fig:compared_with_bicubic_and_srcnn_results}), and legacy cartoons
(``Rangers'' in Fig.~\ref{fig:compared_with_bicubic_and_srcnn_results}).
Note that, we have no access to the original progressive frames (groundtruth) of 
these videos. Without groundtruth, we can only
compare our method to existing methods visually, but not quantitatively.

To evaluate quantitatively (with comparison to the groundtruth), we 
synthesize a set of test interlaced videos from progressive scan videos of different
genres. None of these synthetic interlaced videos exist in our training data.
Fig.~\ref{fig:compared_with_deinterlaced_method} presents a set of synthetic interlaced 
videos, including sports (``Basketball''), scenic
(``Taxi''), computer-rendered (``Roof''), movies (``Jumping''), and
cartoons (``Tide'' and ``Girl''). Due to the page limit, we only present one
representative interlaced frame for each video sequence. While two full size
frames can be recovered from each single interlaced frame, we only show the
first frame in all our results. Please refer to the supplementary materials for
more complete results.

\vspace{0.15in}
\noindent\emph{Visual Comparison}\,\,\,\,
We first compare our method with the classic bicubic interpolation and the
existing DCNN tailored for super-resolution, i.e. SRCNN~\cite{dong2016image}.
Since SRCNN is not designed for deinterlacing, we re-train their model with our
prepared dataset for deinterlacing purpose. The results are presented in
Fig.~\ref{fig:teaser} and~\ref{fig:compared_with_bicubic_and_srcnn_results}. ``Soccer'', ``Bus'' and ``Tennis'' are in $1080i$ format and exhibit severe interlacing
artifacts. Besides, the frames also contain motion-blur and video
compression artifacts. Since both bicubic interpolation and SRCNN reconstruct
each frame from a single field alone, their results are unsatisfactory and
exhibit obvious artifacts due to the large information loss. 
SRCNN performs even worse than the bicubic interpolation, since it
follows the conventional translation-invariant assumption which not held in
deinterlacing scenario. In comparison, our method can obtain much clearer and sharper
results than our competitors. The ``Hunter'' example shows a moving character
from a gameplay where the computer-rendered object
contours/boundaries are sharply preserved. Both bicubic interpolation and SRCNN lead
to blurry and zig-zag near these sharp edges. In contrast, our method obtains
the best reconstruction result in achieving sharp and smooth boundaries. The ``Haystack''
and ``Rangers'' examples are both taken from legacy DVDs in interlaced NTSC
format. In the ``Haystack'' example, only the character is moving, while the
background remains static. Without considering the temporal information, both bicubic
interpolation and SRCNN fails to recover the fine texture of the haystacks and
obtain blurry results. In sharp contrast, our method successfully recovers the
fine texture by taking two fields into consideration.

\begin{table*}
	\center
	\begin{tabular}{|c|c|c|c|c|c|c|}
		\hline
		PSNR/SSIM & Taxi         & Roof & Basketball & Jumping      & Tide  &   Girl \\\hline
		bicubic & 31.56/0.9453          & 33.11/0.9808      &34.67/0.9783          &37.81/0.9801    		 &31.87/0.9809                   & 29.14/0.9585  \\  \hline
		ELA     & 32.47/0.9444          & 34.41/0.9839      &32.08/0.9605             &38.82/0.9844             &33.89/0.9811                 & 31.62/0.9724  \\ \hline
		WLSD    & 35.99/0.9746          & \textbf{35.70}/\textbf{0.9883}      &35.05/0.9794    &38.19/0.9819             &34.17/0.9820                 & 32.00/0.9761 \\ \hline
		FBA     & 34.94/0.9389          & 35.26/0.9815      &33.93/0.9749             &38.27/0.9822             &35.15/\textbf{0.9822}                 & 31.78/0.9756  \\ \hline
		SRCNN   & 30.12/0.9214          & 32.01/0.9749      &29.18/0.9353             &36.81/0.97094             &33.02/0.9758                 & 27.79/0.9477  \\ \hline
		Ours    &\textbf{38.15}/\textbf{0.9834}  & 35.44/0.9866   & \textbf{36.55}/\textbf{0.9838}         &\textbf{39.75}/\textbf{0.9889}    &\textbf{35.37}/0.9807      & \textbf{35.44}/\textbf{0.9866} \\ \hline
	\end{tabular}
	\caption{PSNR and SSIM between the deinterlaced frames and groundtruth of all methods.}
	\label{tab:psnr_comparsion}
\end{table*}

\begin{table*}[!tp]
	\center
	\begin{tabular}{|c|c|c|c|c|c|c|c|}\hline
		\multirow{2}{*}{Average time (s)} & \multirow{2}{*}{ELA} &\multirow{2}{*}{WLSD} &\multirow{2}{*}{FBA} & \multirow{2}{*}{Bicubic} & \multirow{2}{*}{SRCNN} &  \multicolumn{2}{c|}{Our Methods} \\ \cline{7-8}
		& & & & & & With sharable layers & Without sharable layers\\ \hline
		$1920\times 1080$ & 0.6854  &2.9843  &4.1486 &0.7068 &0.3010 & \textbf{0.0835} & 0.2520\\ \hline
		$1024\times 768$  & 0.0676  &1.0643  &1.6347 &0.2812 &0.0998 & \textbf{0.0301} & 0.0833\\ \hline
		$720\times 576$   & 0.0317  &0.4934  &0.6956 &0.1176 &0.0423 & \textbf{0.0204} & 0.0556\\ \hline
		$720\times 480$   & 0.0241  &0.4956  &0.7096 &0.1110 &0.0419 & \textbf{0.0137} & 0.0403\\ \hline
	\end{tabular}
	\caption{Timing statistics for all methods.}
	\label{tab:time_statistics}
\end{table*}

We further compare our method to the state-of-the-art deinterlacing methods,
including ELA~\cite{doyle1990interlaced}, WLSD~\cite{wang2014interlacing}, and
FBA~\cite{vedadi2013interlacing}. ELA is the most widely used deinterlacing
methods due to its high performance. It is an intra-field method and uses edge
directional correlation to reconstruct the missing scanlines. WLSD is the state-of-the-art intra-field deinterlacing method based on optimization. It generally
produces better result than that of ELA, but with a higher computational
expense. FBA is the state-of-the-art inter-field method.
Fig.~\ref{fig:compared_with_deinterlaced_method} shows the results of all
methods for a set of synthetic interlaced videos, in which we have the
groundtruths for quantitative evaluation. Besides the reconstructed frames, we
also blow-up the difference images for better visualization. The difference
image is simply computed as the pixel-wise absolute difference between the
output and the groundtruth. As we can observe, all our competitors
generate artifacts surrounding the boundaries. The sharper the boundary is, the more
obvious the artifact is. In general, ELA produces the most artifacts since it
adopts a simple interpolator and utilizes information from a single field alone. 
WLSD produces less artifacts as it adopts a more complex optimization-based strategy to fill the missing pixels. But it still
only utilizes information of a single field and has large information loss
during reconstruction. Though FBA utilizes the temporal information, 
it still cannot achieve good visual quality because
they only rely on simple interpolators. In contrast, our method produces
significantly less artifacts than all competitors.

\begin{figure}[!tp]
	\centering
	\includegraphics[width=0.9\linewidth]{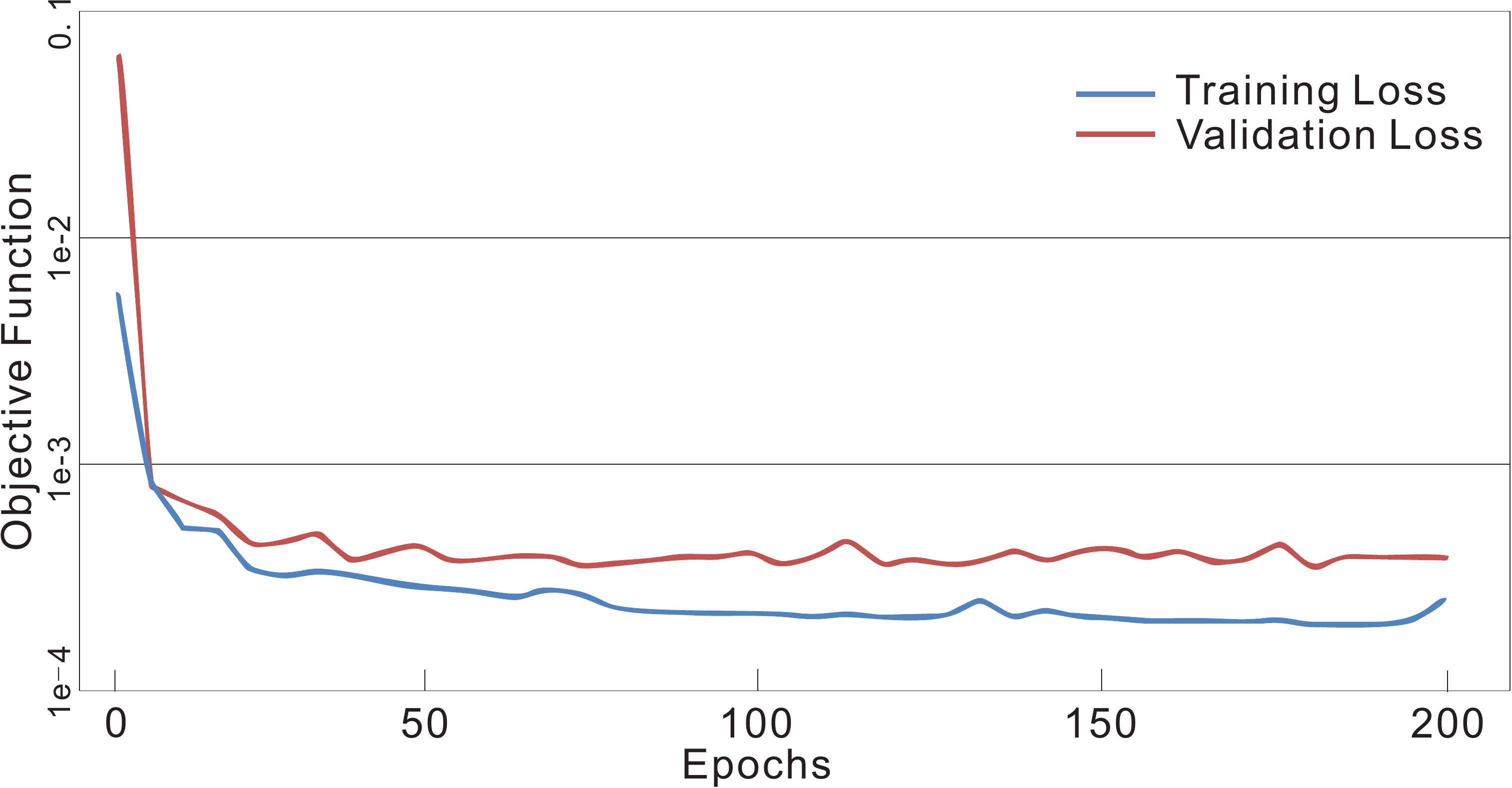}\\
	\caption{Training loss and validation loss of our neural network.}\label{fig:training_loss}
\end{figure}

\vspace{0.15in}
\noindent\emph{Quantitative Evaluation}\,\,\,\,
We train our neural network by minimizing the loss of Eq.~\ref{eq:objective} on
the training data. The training loss and validation loss throughout the whole
training epochs are shown in Fig.~\ref{fig:training_loss}. Both training and
validation losses reduce rapidly after the first few epochs and converge in
around 50 epochs.

We also compare the accuracy of our method to our competitors in terms of peak
signal-to-noise ratio (PSNR) and structural similarity index (SSIM). Note that
we only compute the PSNR and SSIM for those test videos with groundtruth. We
take the average value over all frames of each video sequence in computing both
measurements. Table~\ref{tab:psnr_comparsion} presents the statistics. Our
method outperforms the competitors in terms of both PSNR
and SSIM in most cases.

\vspace{0.15in}
\noindent\emph{Timing Statistics}\,\,\,\,
Lastly, we compare the running time of our method to our competitors on a
workstation with Intel Core CPU i7-5930, 65GB RAM equipped with a nVidia TITAN X
Maxwell GPU. The statistics are presented in Table~\ref{tab:time_statistics}.
Our method achieves the highest performance among all methods in all
resolutions. It processes even faster than ELA with apparently better visual
quality. ELA and SRCNN have similar performance and are slighter slower than our
method. Bicubic interpolation, WLSD, and FBA have much higher computational
complexity and are far from real-time processing.  Note that ELA is only a CPU
method without GPU acceleration. In particular, with a single GPU, our method
already achieves real-time performance up to the resolution of $1024\times 768$
(33 fps). With one more GPU, our method can also achieve real-time performance
for $1920\times 1080$-resolution videos. We also test our model without sharing
lower-level layers, i.e., two separate networks are needed for reconstructing
the two frames. The statistics is shown in the last column in
Table~\ref{tab:time_statistics}. This strategy roughly triples the computational
time while quality is similar to that with sharing low-level layers.


\begin{figure}[!tp]
	\centering
	\includegraphics[width=\linewidth]{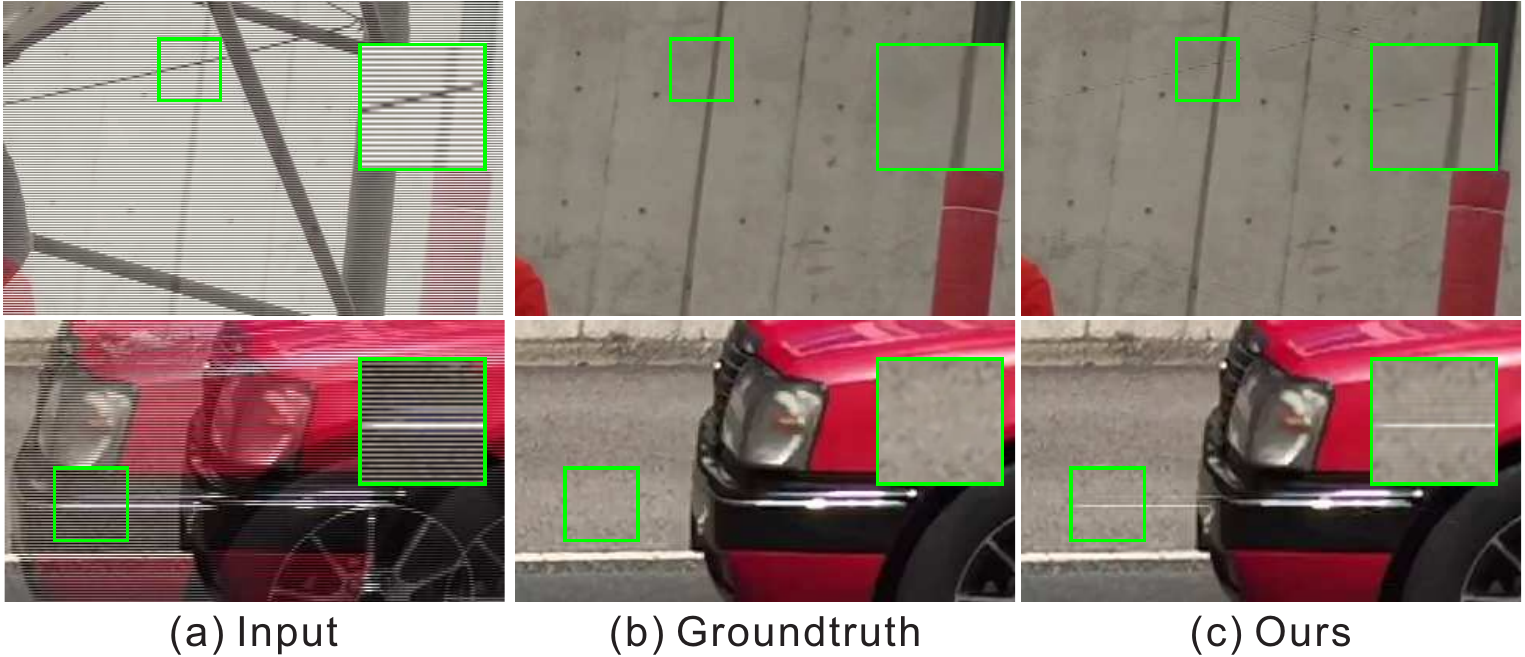}\\
	\caption{Failure cases. The top row shows a case where our result contains obvious artifacts when the motion of the interlaced frame is too large. The bottom row shows a case where our method fails to identify thin horizontal structures as interlacing artifacts and incorrectly preserves it in the reconstructed frame.}\label{fig:failure_cases}
\end{figure}

\vspace{0.15in}
\noindent\emph{Limitations}\,\,\,\,
Since our method does not explicitly separate the two fields for reconstructing
two full frames, the two fields may interfere each other 
badly when the motion between the two fields are extremely large. The first row
in Fig.~\ref{fig:failure_cases} presents an example where the interlaced frame
has a very large motion, obvious artifacts can be
observed. Our method may also fail when the interlaced frame contains very thin
horizontal structures. The second row of Fig.~\ref{fig:failure_cases} shows an
example where a horizontal thin reflection stripe appears on a car. Only one line
of the reflection stripe is scanned in the interlaced frame. Our neural network
fails to identify it as a result of interlacing, but regards it as the original
structures and incorrectly preserves it in the reconstructed frame. This is
because this kind of patches is rare and gets diluted by the large amount of
common cases. We may relieve this problem by training the neural network with 
more such training patches. 

%% file: conclusion.tex
In this paper, we present the first DCNN for video deinterlacing. Unlike
the conventional DCNNs suffering from the translation-invariant issue, we
proposed a novel DCNN architecture by adopting the whole interlaced frame as
input and two half frames as output. We also propose to share the lower-level
convolutional layers for reconstructing the two output frames to boost  
efficiency. With this strategy, our method achieves real-time deinterlacing
on a single GPU for videos of resolution up to $1024\times768$.
Experiments show that our method outperforms existing methods,
including traditional deinterlacing methods and DCNN-based models re-trained for
deinterlacing, in terms of both reconstruction accuracy and computational
performance.

Since our method takes the whole interlaced frame as the input, frame
reconstruction is always influenced by both fields. While this may produce
better results in most of the cases, it occasionally leads to visually poorer
results when the
motion between two fields is extremely large. In this scenario, reconstructing
each frame from a single field without considering temporal information may
produce better results. A possible solution is to first recognize such large-motion
frames, and then decide whether temporal information should be utilized for 
deinterlacing.
